# Tipping the Scales: A Corpus-Based Reconstruction of Adjective Scales in the McGill Pain Questionnaire

Miriam Stern

Program in Linguistics, Princeton University, Princeton, New Jersey
`mestern@princeton.edu`

## Abstract

*Modern medical diagnosis relies on precise pain assessment tools in translating clinical information from patient to physician. The McGill Pain Questionnaire (MPQ) is a clinical pain assessment technique that utilizes 78 adjectives of different intensities in 20 categories to quantify a patient's pain. The questionnaire's efficacy depends on a predictable pattern of adjective use by patients experiencing pain. In this study, I recreate the MPQ's adjective intensity orderings using data gathered from patient forums and modern NLP techniques. I extract adjective intensity relationships by searching for key linguistic contexts, and then combine the relationship information to form robust adjective scales. Of 17 adjective relationships predicted by this research, only 4 diverge from the MPQ's orderings, which is statistically significant at the 0.1 alpha level. The results suggest predictable patterns of adjective use by people experiencing pain, but call into question the MPQ's categories for grouping adjectives.*

## 1. Introduction

The question of pain's communicability is a crucial one; to treat pain, it must first be identified and categorized. In the past few decades, the practice of using numbers to describe pain has been questioned by linguists, medical professionals, and others [1],[2],[3]. Specifically, studies have shown that clinical data that takes only pain intensity into consideration is insufficient [1],[3]. This paper considers the McGill Pain Questionnaire (MPQ), the most commonly used verbal pain assessment tool that asks patients to describe their pain using a provided list of adjectives [1],[4]. In collecting a combination of adjectives, the MPQ is meant to extract more nuanced information than can a numerical rating system that considers only intensity data [1][2].

The MPQ has several sections, but this research focuses on the one entitled "What Does Your Pain Feel Like?" This section asks patients to select up to one word in each of 20 adjective categories that describes their pain. Each category contains between two and six words, which the MPQ posits are gradations of the same sensation, and which are assigned a numerical value accordingly. For instance, category 11's "tiring" and "exhausting" are different intensities of one scalar property, with "tiring" assigned a value of 1 and "exhausting" a value of 2. In this way, the MPQ seeks to translate qualitative verbal descriptions into quantitative numerical data to communicate pain [1].

An analysis of the body of literature on pain assessment tools reveals that various components of the MPQ have been independently retested by other researchers since its construction [5]-[8]. Several studies have shown that the MPQ generally has good construct validity [6]-[8], however, one study called into question the correctness of the MPQ's adjective groupings, suggesting that the adjectives might be imprecisely categorized [5]. There are additional weaknesses revealed in the studies conducted to construct and then verify the MPQ. Firstly, the original MPQ research used mixed populations of doctors and patients to construct the adjective intensity scales, with the two groups being observed to assign different values to pain adjectives [2]. Since the goal of pain assessment questionnaires is to facilitate the communication of a patient's pain, the ways in which patients actually communicate about their pain is vitally important. Additionally, all studies conducted on the MPQ's efficacy have been laboratory experiments, with the adjectives provided to the patients by the researchers. Thus, the patients' interactions with the adjectives were not entirely natural [5]-[8]. Finally, since the MPQ was constructed in the 1970's, and language is constantly evolving, the time-dependency of the MPQ's adjectives must also be called into question.

My research improves upon prior methods by analyzing corpus data drawn from self-authored forum postings of pain sufferers. To assess the MPQ's system of ranking adjectives by intensity, I created a text corpus from online chronic pain forum posts. Using this corpus, I conducted a search for specific linguistic contexts in which adjectives are used; from these contexts, the intensity relationships amongst pairs of adjectives was inferred. By combining these generated intensity relationships, I constructed novel adjective intensity scales and compared them to the ones present in the MPQ. This approach allows for the analysis of spontaneously produced pain descriptions, which reflect each author's pain most personally. Furthermore, the contemporariness of the blog postings provided insights into the timelessness or lack thereof of the adjective scales devised by the MPQ's creators Finally, the construction of these adjective scales requires that patients use adjectives in a consistent manner to describe their pain. As such, successful reconstruction of adjective intensity scales would validate the concept behind using an adjective questionnaire to elicit medical and diagnostic data.

As a preliminary study, this data corroborates the adjective intensity scales defined by the MPQ at above chance levels. This research suggests, however, that the categories for the MPQ adjectives may be imprecisely defined. Accordingly, further research will be required to fully analyze the validity of the MPQ.

## 2. LINGUISTIC BACKGROUND

To understand how verbal pain questionnaires work, a discussion of adjective scales in general is needed. In this section, I will introduce the concept of "scalar implicature", and then consider its role in adjective meaning and intensity.

### 2.1. Introduction to Scalar Implicature

There is a distinction drawn in linguistics between meaning that is conveyed explicitly or verbally and meaning that is implicitly derived. For example, consider the following statement:

1.
    a. Ezekiel likes some of the teachers in his school.

At surface level, there is a literal meaning conveyed by the words of this statement. However, in any world in which (1a) is true, (1b) must also be true. This is an example of *entailment*, which is meaning conveyed automatically and immutably by a statement. Additionally, someone who heard statement (1a) would typically understand it to imply (1c) [9].

    b. Ezekiel likes at least one teacher in his school.
    c. Ezekiel does not like all of the teachers in his school.

The relationship between (1a) and (1c) is an example of *implicature*, which refers to meaning that is implied by an utterance without being explicitly stated. Unlike (1b), the statement made by (1c) can be cancelled, as demonstrated by (1d), or reinforced, as demonstrated by (1e) [9],[10].

d. Ezekiel likes some of the teachers in his school; in fact, he likes them all.
e. Ezekiel likes some of the teachers in his school, but not all of them.

Though there are different types of implicature, this example demonstrates *scalar implicature*: involving words like 'some' and 'all,' scalar implicature is concerned with the ways in which word meanings differ in intensity, and the additional information that they can convey [9],[10]. In this case, conversational conventions dictate that if the speaker knew that Ezekiel likes *all* his teachers, the speaker would be expected to say that Ezekiel likes *all* of his teachers (since *all* is a more informative statement than *some*). The fact that the speaker said *some* and not *all* therefore implies that the speaker was not able to say *all* [11].

This concept of scalar implicature applies to other adjectives as well, which can be organized in "Horn scales" [9],[10]. As an example, on the scale <pretty, beautiful> *beautiful* entails *at least pretty*, though *pretty* does not entail *at least beautiful*. Thus, a term on the left of a pair like <pretty, beautiful> suggests that any term to the right is inapplicable, or at least not known to be applicable [10],[12].

These properties of scalar implicature and adjective scales form the basis of adjectival pain assessment tools. In particular, adjectival pain assessment tools rely on the assumption that, if a patient describes her pain using adjective A from among a particular scale, the patient is describing her pain as A, *to the exclusion of any stronger description* in the same category [1].

## 3. HYPOTHESIS

Building upon the body of literature surrounding adjective scales and pain questionnaires, I evaluated the adjective groupings presented by the MPQ. Specifically, I attempted to assemble Horn scales using the adjectives found in the MPQ by analyzing online forum data. Comparing these constructed scales to the categorical hierarchies prescribed by the MPQ provided insight into the ways in which people use adjectives to describe their pain, and thus into the MPQ's validity. This research sought to reject the null hypothesis $H0_a$ and affirm the hypothesis H1.

> $H0_a$: There is no predictable pattern to the way in which people use scalable adjectives.

> H1: There is a predictable pattern to the way in which people use scalable adjectives.

Additionally, using unique chronic pain forums dedicated to specific types of pain, this research considered whether people suffering from different diseases or ailments tend to use unique frequency distributions of adjectives to describe their pain. This leads to another set of hypotheses:

> $H0_b$: Different categories of chronic pain are not associated with specific adjective pain descriptors.

> H2: Different categories of chronic pain are associated with specific adjective pain descriptors.

The efficacy of the MPQ and other similar metrics depends on the predictable and unambiguous way in which scalable adjectives are utilized; an adjective description of pain is only useful insofar as its meaning is mutually agreed upon by the patient providing the adjective and the physician receiving it. Therefore, affirmation or rejection of these hypotheses provides useful insights into the rationality of using an adjectival pain scale for clinical diagnoses.

## 4. CORPUS

### 4.1. Web Scraping and Pre-Processing

For the textual data needed in this research, I created a corpus of text produced by patients experiencing pain. I chose internet forums as the source of this data because of their public and voluntary nature, and the wide range and specificity of forum topics. Crucially for this research, forum postings reflect speech occurring naturally amongst patients, rather than between patients and physicians.

For this paper, data was pulled from a website called HealingWell.com, which is meant to be an online community for those experiencing chronic pain [13]. Within this website, data was pulled from forums of the following three topics: Chronic Pain, Rheumatoid Arthritis (RA), and Fibromyalgia. Chronic Pain was chosen as a catch-all for the different types of pain that people experience. RA and Fibromyalgia were both chosen as conditions for which pain is a primary symptom, according to their Mayo Clinic descriptions [14],[15]. I selected these three categories to consider whether there might be a different distribution of adjective use in pain descriptions between the different categories—RA and Fibromyalgia—and between the two categories and the broader Chronic Pain forum.

From the HealingWell website, I scraped data from the three forums of interest [13],[16]. Specifically, I collected all the text from blog postings on each of the three topics. Then, I preprocessed the data by tokenizing the text [17], and correcting typos [18].

### 4.2. Corpus Description

Among the three forums (Chronic Pain, RA, and Fibromyalgia), the Chronic Pain forum was the largest, with a total of 20,189,291 words. Rheumatoid Arthritis was the second largest, with 4,160,952 words, and the smallest of the three was the Fibromyalgia forum, with 4,156,802 words. Since adjectives are of interest in this paper, I calculated additional statistics for the adjectives in each forum text. Basic statistics about the forum texts are summarized in Table 1.

Table 1. Forum Data Descriptions.

|  | Chronic Pain | Rheumatoid Arthritis | Fibromyalgia |
|---|---|---|---|
| Total Words | 20,189,291 | 4,160,952 | 4,156,802 |
| Mean Post Length (words) | 145 | 123 | 117 |
| Median Post Length (words) | 101 | 89 | 75 |
| Post Length Range | 1, 3681 | 1, 3456 | 1, 5047 |
| Unique Tokens | 193,504 | 43,733 | 73,561 |
| Type/Token Ratio (%) | 1.770 | 1.051 | 0.9584 |
| Total Adjectives | 343,665 | 331,462 | 1,959,734 |
| Unique Adjectives | 14,486 | 14,603 | 34,196 |

### 4.3. Linguistic Processing

Once the initial data collection and processing was complete, I categorized and tagged the textual data using an off-the-shelf part-of-speech (POS) tagger [17]. An abridged list of parts of speech with their corresponding tag abbreviation is given in Table 2 [17].

Since the primary part of speech of interest in this paper is the adjective ('JJ'), I gave special care to these tags to ensure their proper identification. An example sentence from the data is given below in the form 'word/TAG'.

(1) 'the/DT pain/NN is/VBZ a/DT different/JJ pain/NN ,/, the/DT best/JJS way/NN i/NN can/MD describe/VB it/PRP is/VBZ a/DT deep/JJ burning/NN pain/NN ,/, not/RB a/DT neuropathy/JJ pain/NN either/DT ./.'

As can be seen from the example sentence, the default NLTK POS-tagger is not 100% accurate: here, the word 'burning' would be more accurately categorized as an adjective. I will address this problem in the next section.

Table 2. Select Part of Speech Tagset.

| Tag | Part of Speech |
|-----|----------------|
| CC  | Coordinating conjunction |
| DT  | Determiner |
| IN  | Preposition/subordinating conjunction |
| JJ  | Adjective |
| JJR | Adjective, comparative |
| JJS | Adjective, superlative |
| MD  | Modal could, will |
| NN  | Noun, singular |
| NNS | Noun, plural |
| RB  | Adverb |
| VB  | Verb |

### 4.3.1 Special Considerations

In English, certain suffixes are associated with specific types of speech. For example, if a word is tagged with the suffix [-s], as in the word 'cat-s', that word can be assumed to be a noun, since nouns take the plural suffix [-s]. However, English has many ambiguous morphemes, or meaning-carrying units, which undermine generalizations like the one introduced for the [-s] suffix. For example, words ending with [-ing] might be categorized as a verb, noun, or adjective. Examples of this are given in sentences (2)-(4), using the English root 'burn.'

(2) The ceremonial *burning* of the torch takes place tomorrow.
(3) The *burning* building could be seen from miles away.
(4) A light was *burning* in the hallway.

'Burning' assumes the role of noun (2), adjective (3), and verb (4) in these three sentences, with its meaning parsed based on the syntactic and semantic position of the word in the sentence. For this research, it was especially important that adjectives like 'burning' not be categorized as verbs. As such, I added an additional condition to the pos-tagger, such that all words ending in '-ing' that immediately preceded a noun were tagged as adjectives. This was meant to ensure that a usage such as (3) would be appropriately labeled as ADJ, whereas 'burning' in (2) and (4) would be unaffected.

In the final stages of data analysis, I still found instances where MPQ adjectives were mislabeled as nouns specifically. To guarantee that all instances of MPQ adjectives would be collected, the final step in tailoring the POS-tagger consisted of feeding the tagger a list of all the adjectives from the MPQ; any time the adjective appeared in the right morphological form, the automatic tag was overridden with the adjective tag.

### 4.4. Corpus Comparison

I assembled the corpus used for this research from specifically oriented text: all the forum postings were on the topic of chronic pain. To understand how broadly applicable the results of this research are, it is important to consider how the makeup of this corpus compares to other more topic-neutral corpora. I selected two other corpora for this comparison. The Brown Corpus is a roughly million-word corpus compiled from different genres of texts published in English in 1961 [19]. Though the Brown corpus is small and somewhat outdated, I opted to include it in the comparison since it was released close to the date of the MPQ's creation; if there is any time-dependence on the frequency of adjective use, the Brown corpus might show divergence from the other corpora.

I also included for comparison the much larger, billion-word Corpus of Contemporary American English (COCA). Comprised of spoken and written text collected from 1990 through the present day, this corpus is meant to serve as a big-picture sample of American English [20]. I collected frequency data of each MPQ adjective in each of the three corpora.

After assembling the frequency counts of each MPQ adjective in each of the corpora, I calculated a cosine similarity rating for each pair. The cosine similarity ratings between the corpora were as follows: HealingWell and Brown: 0.794; HealingWell and COCA: 0.620; Brown and COCA: 0.776. The cosine similarity value ranges between 0 and 1, with 0 indicating no overlap between the two sets, and 1 indicating complete agreement [21]. Since the three cosine similarity ratings are all high, there does not appear to be significant or categorical disparities in MPQ adjective frequency across different corpora or different contexts.

A similar analysis was done within the HealingWell corpus itself in relation to the different forum topics considered. For each of the forum topics—RA, Fibromyalgia, and Chronic Pain—the frequency of the MPQ adjectives was calculated. Graphing the data, the overall patterning of adjective frequency was consistent across forum topics. RA appears to be somewhat of an outlier for a few adjectives, including 'sore', 'itchy', and 'aching', for which the RA frequency is significantly higher than the other two topics.

## 5. METHODOLOGY

### 5.1. Identifying Adjective Contexts

After creating a corpus, I conducted searches to find the adjective contexts needed to construct adjective intensity scales. This section outlines the process for defining and locating the linguistic contexts of interest.

The target for this search was the type of construction described by Horn, whereby the pattern of a sentence containing two adjectives can convey the intensity relationship between the two. Examples of Horn's adjective constructions are provided in (5)-(7) where adjective Y has a stronger intensity than adjective X [10]. For each construction form, an example sentence is provided to its right.

(5) X but not Y // Warm but not hot
(6) Not only X, but Y // Not only ugly, but grotesque
(7) X if not Y // Bright if not blinding

In addition to these examples from Horn, I also gathered intensity patterns by reviewing Sheinman et al. I utilized Sheinman's strategy of breaking down the examples into two categories: intense patterns and mild patterns. Intense patterns contain two adjectives X and Y such that Y is more intense than X [22]. An example of an intense pattern is given in (8).

(8) X, perhaps even Y // Good, perhaps even great

Mild patterns contain two adjectives X and Y such that X is more intense than Y. An example of an intense pattern is given in (9).

(9) Not Y but still very X // Not ginormous but still very big

I created a new list of each construction type using examples from Horn, Sheinman et al., and several novel patterns. The final list of intense and mild patterns totaled 13, with 6 intense and 7 mild patterns. I lay out the patterns of each type in Table 3.

I analyzed the constructed corpus data in 10-word chunks to search for matches against any of the specified mild and intense patterns. Though the intense and mild patterns are all 6 words or fewer, I opted to use n-grams with n=10 and to expand the regular expressions to allow for additional complexity in the phrases matched. That is, while the sentence 'all day at work it is

**stiff but not painful**' would return a match using an n-gram with n=5, the phrase 'the pain is quite **sharp but not** usually particularly **achy**' would not. Using my approach, both phrases returned matches.

To further refine the search, I also ran the matched phrases against the list of MPQ adjectives. Phrases which contained two or more adjectives (at least one on either side of the intense/mild construction) from the MPQ list were selected, while all others were excluded. After this stage, I collected a total of 114 phrases: 27 from RA; 38 from Fibromyalgia; and 49 from Chronic Pain.

Table 3. Intense and Mild Patterns

| Intense Pattern | Mild Pattern |
| --- | --- |
| if not X at least Y | X but not Y |
| not X but Y enough | X but never Y |
| not X just/only Y | X but hardly Y |
| not X but still (very) Y | X even/perhaps Y |
| not/no X just/only Y | X perhaps/and even Y |
| no X just Y | X almost/if not/sometimes Y |
|  | X sometimes almost/even Y |

### 5.2.1. Excluded Data

Some phrases that matched the listed criteria were not suitable for analysis in this research. From the original 114 matching phrases, I discarded an additional 46 as 'false positives.' I excluded these phrases due to three remaining issues.

*Wrong Topic*. First, not all the identified phrases were on-topic; that is, adjective constructions that fit all the other criteria were sometimes describing non-pain-related subjects. Examples of phrases that were discarded for this reason are given in (10) and (11).

(10) you/prp melt/vbp the/dt wax/nn so/in its/prp$ **hot/jj but/cc not/rb burning/jj**

(11) i/prp love/vbp swimming/vbg but/cc **cold/jj** water/nn or/cc **even/rb cool/jj** water/n

Off-topic phrases represented the largest subset of rejected matches, with a total of 35 off-topic phrases discarded between the three forums.

*Wrong Tag.* Secondly, as mentioned previously, POS-taggers often have difficulty distinguishing between gerunds ('VBZ') and modifiers ('JJ'). To ensure that all MPQ adjectives were accounted for, the data processing included overriding all tags of MPQ adjective roots and replacing them with the adjective tag 'JJ'. This default adjective tagging favored false positives over false negatives. As such, certain words suffixed with [-ing] were tagged as adjectives, even while being used as verbs or others in their contexts. Examples of phrases that were excluded for having mis-tagged adjectives are given in (12) and (13).

(12) **tight/jj** and/cc my/prp$ heart/nn **sometimes/rb** feels/nns like/in it/prp is/vbz **beating/jj**

(13) **tingling/jj** as/in **perhaps/rb** it/prp could/md be/vb **pressing/jj** on/in a/dt nerve/nn

Of 6 total phrases discarded for mis-tagging, four different words were mis-tagged: 'killing' (2), 'beating' (2), 'pressing' (1), and 'cutting' (1).

***Wrong Noun.*** Finally, some strings that matched the specified patterns contained multiple phrases in one sentence. Examples of this are given in (14) and (15).

(14) so/rb **annoying/jj** yes/uh **itchy/jj sometimes/rb** a/dt **hot/jj** soak/nn helps/vbz ./.

(15) not/rb suggest/vb just/rb using/vbg **cold/jj** pools/nns or/cc **even/rb hot/jj** tubs/nns

In (14), the adjectives 'annoying' and 'itchy' describe a sensation, whereas 'hot' modifies the noun 'soak.' Here, the issue is one of missing punctuation, which is common in internet writing. However, in (15), there are two clauses within one sentence separated by a coordinating conjunction, such that adjectives on either side of the conjunction modify two separate nouns. Since the two adjectives are being used separately and cannot readily be ranked on one intensity scale, an intensity relationship cannot be inferred. There were 5 total phrases rejected for including separately describing adjectives.

After I sorted through all the data and removed false positives, a total of 66 matched phrases remained for use in data analysis.

## 5.2. Adjective Scale Construction

Once I compiled and searched the corpus, I then used the data collected to address the research question pursued in this paper. In this section, I present the process of using the partially ordered weak-strong pairs discussed in Section 5.1 to construct adjective intensity scales.

### 5.2.1. Weak-Strong Pairs

Using the final list of matched phrases, each phrase was analyzed to yield one, or more, weak-strong adjective pair. First, I divided the phrases by the type of pattern they had matched: mild or intense. Within each group, I identified the conjunction string for each phrase based on the pattern matched. As an example, if a phrase matched the mild pattern 'x and even y,' the conjunction string would be 'and even.' For all mild patterns, I assigned the adjective on the left of the conjunction string the value *weak*, and the adjective on the right the value *strong*. If instead a phrase matched an intense pattern, I assigned the left adjective the value *strong* and the right adjective the value *weak*.

Where a phrase contained multiple adjectives on either side of the conjunction string, the same process was applied for each of the adjectives on either side of the construction, as shown using the example in (16).

(16) back/nn areas/nns can/md feel/vb **heavy/jj aching/jj** and/cc yes/jj <u>sometimes</u>/rb **burning/jj**

Here, there would be two strong-weak pairs identified—heavy-burning, and aching-burning—though there is no limit on the number of pairs that could be derived from one phrase.

In all, I identified 81 weak-strong pairs. Of the 81, 17 were found in the RA text, 25 in Fibromyalgia, and 39 in Chronic Pain. I then ran this list of pairs through a lemmatizer [17]. For the data at hand, this meant that instances of 'itchy' and 'itching' were identified with the same root of 'itch.' Similarly, 'achy' and 'aching' were identified with the same root of 'ache.' Though there are nuanced differences in meaning between the separately inflected forms of the same root, for the purposes of this preliminary analysis, this treatment of different inflectional forms was sufficient. After lemmatization, 27 of the MPQ's 78 total adjectives were accounted for in at least one of the weak-strong pairs. The adjectives occurring in the greatest number of pairs were 'burning' (20), 'sharp' (19), 'tingling' (14), and 'aching' (13).

### 5.2.2. Adjective Categorization

Once I identified the weak-strong pairs, I then combined the partially ordered pairs to make categorical adjectival scales.

As mentioned previously, the MPQ divides its adjectives into 20 different categories. This organizational structure depends on the idea that, within each category, the adjectives all describe the same sort of sensation, differing only in intensity [1],[2]. Since our research's list of 81 weak-strong pairs was derived from spontaneous speech, the two adjectives in each pair were not necessarily found in the same MPQ category. For example, consider the following sentence in example (17).

(17)    no **hurting** or **prickling** feelings just the **numb tingling** feeling

In this sentence, there were four weak-strong pairs identified which involve adjectives from four different MPQ categories. The pairs, given in the form (weak: category, strong: category) are summarized as follows:

(18)    numb:18, hurting:9
numb:18, prickling:3
tingling:8, hurting:9
tingling:8, prickling:3

Though there are clearly some differences in intensity between the adjectives, as demonstrated by the intense pattern 'no x, just y', the MPQ does not attempt to put adjectives of different categories on the same scale. This can be illustrated more clearly using adjectives that are not exclusively used to describe pain: cold/freezing and hot/burning. Though it is intuitively clear that 'freezing' is more intense than 'hot', and 'burning' more intense than 'cold,' the elements cannot all be incorporated into a Horn scale that maintains the principles of implicature [9],[10]. An example of a possible (19) and impossible (20) scale are given in the following examples:

(19)    <cool, cold>
(20)    *<cold, cool, warm, hot>

Using Horn's concepts of scalar implicature, if the scale given in (20) were possible, we would expect *cool* to implicate *not warm* just as *warm* implicates *not hot* [10]. Given the question "How is the temperature of your tea?" we would expect both (21) and (22) to be acceptable answers, where the implicature is expressed explicitly.

(21)    It's warm, but not hot.
(22)    *It's cool, but not warm.

Given that (22) is an unacceptable sentence, the scale provided in (20) is an impossible one; hence, I limited the analysis for this research to adjectives within the same MPQ category. However, since not all weak-strong pairs identified in the data involved adjectives of the same category, the next step in the process took advantage of transitive relationships between adjectives, as I will describe in the next section.

### 5.3. Graph Creation and Traversal

Constructing robust scales from the weak-strong pairs required a two-step process. First, I plotted all the weak-strong pairs involving MPQ adjectives on a graph. To construct the graph, a node was created for each adjective represented in the MPQ weak-strong pairs. For each pair, an edge was drawn between the two adjectives' nodes, with an arrow pointing from the weaker adjective to the stronger adjective [23],[24]. This graph is presented in Figure 1.

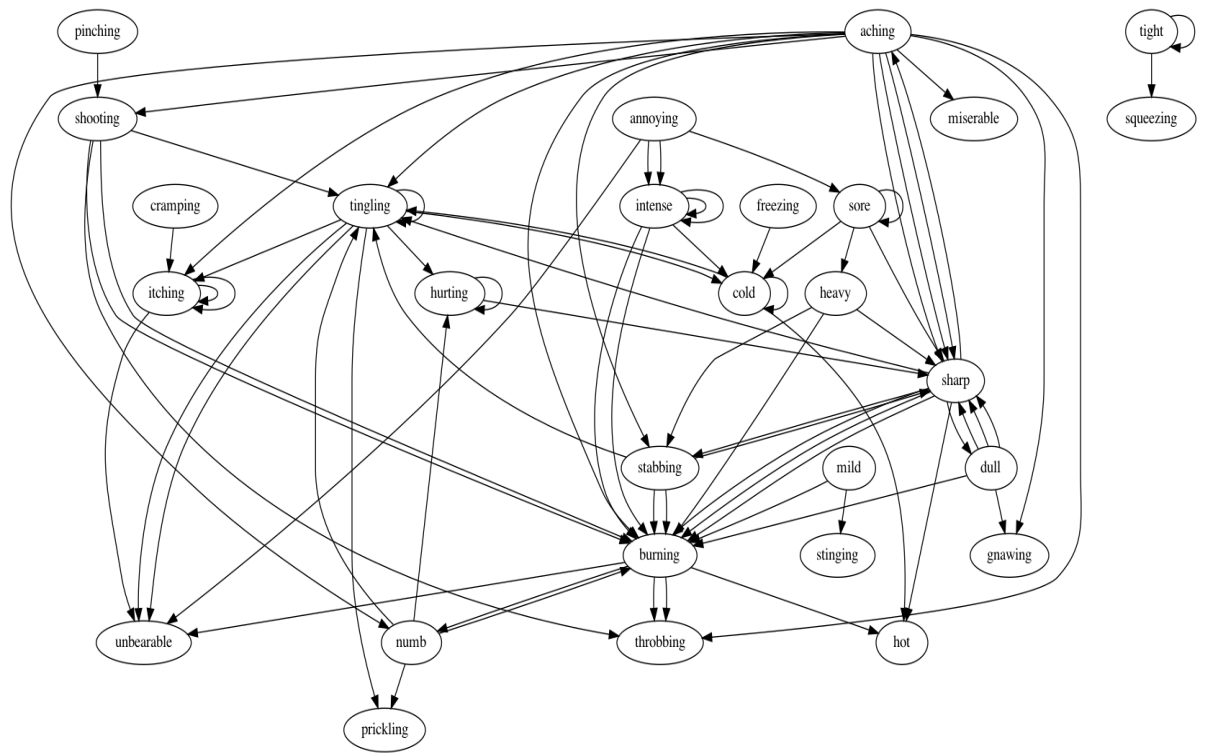

Figure 1. Weak-Strong Adjective Relation Graph

Where multiple arrows begin or end at a particular adjective node, that adjective was present in multiple weak-strong pairs. Several adjective pairs have bidirectional arrows between them. For example, in the case of 'sharp' and 'dull,' there are three arrows pointing from 'dull' to 'sharp,' suggesting that 'sharp' is stronger than 'dull,' and one arrow pointing from 'sharp' to 'dull,' suggesting that 'dull' is stronger than 'sharp.' These arrows are contradictory, as 'sharp' cannot be both stronger and weaker than 'dull.' Intuitively, 'sharp' is in fact stronger than 'dull,' which is reflected in the majority of the weak-strong pairs. In these cases, where arrows connecting two adjective nodes pointed in both directions, I prioritized the majority direction. I will explore this phenomenon, along with the rest of the graph, in more detail in the following data analysis section.

The second step of the scale construction process required graph traversal to find paths between adjectives of the same MPQ category. Ideally, each adjective in each MPQ category would have been found in a mild or intense pattern with another adjective of that same category. However, in the mined dataset, there were only eight instances where two adjectives of the same MPQ category were found in the same weak-strong pair. I present an example of this in (23), where both 'annoying' and 'unbearable' are category 16 adjectives.

(23)    its an annoying pain but not unbearable pain its been

Because of this limited pool of data, I also considered transitive relationships between adjectives of the same MPQ category in the graph traversal. That is, given the two category 9 adjectives 'aching' and 'hurting', if 'aching' is weaker than 'tingling' and 'tingling' is weaker than 'hurting,' then 'aching' must also be weaker than 'hurting.' With transitivity, I was able to deduce intensity relationships even where there was not a direct connection between two adjectives of the same category. I traversed the graph using a recursive search of the graph's nodes, given an input of every possible adjective pair combination within each MPQ category. Starting at the node of one of the adjectives, each path away from the adjective was travelled from node to node until either a) there were no available and unvisited nodes to visit from the current node or b) an adjective of the same MPQ category was found. Using this recursive search, 404 total paths were traced between adjectives of the same MPQ categories, with 22 unique adjective pairs connected.

### 5.4. Scale Construction

After collecting the paths between adjectives and constructing the weak-strong pairs within each MPQ category, I combined the partial orderings to make robust adjective scales for each category. I considered each MPQ category independently. For each weak-strong pair in a given category, I assigned the weak adjective a value of '0' and the strong adjective a value of '1.' Then, I summed together the values for each adjective in a given category. I present an example of this process using category 16 adjectives (24).

(24) Weak-Strong Pair: <annoying, intense>
1. Annoying + 0
2. Intense + 1

Weak-Strong Pair: <annoying, unbearable>
1. Annoying + 0
2. Unbearable: +1

Weak-Strong Pair: <intense, unbearable>
1. Intense: +0
2. Unbearable: +1

Total Values:
Annoying: 0
Intense: 1
Unbearable: 2

Once I calculated the total value for each adjective, I constructed a scale that ranked the intensity relationships of the adjectives in each category in numerical order. For the example in (24), the category 16 adjectives would be ordered on the scale <annoying, intense, unbearable>, in ascending order of intensity.

The final constructed adjective scales for each of the categories represented in the data are presented in Table 2 below. Where two adjectives are separated by a slash, the data was inconclusive on the intensity relation between the two.

Table 2. MPQ Category Scale Reconstruction.

| MPQ Category | Adjective Scale |
| --- | --- |
| 3 | <stabbing, prickling> |
| 7 | <burning, hot> |
| 8 | <tingling, itching> |
| 9 | <dull/sore/aching, heavy, hurting> |
| 16 | <annoying, intense, unbearable> |
| 18 | <tight, squeezing> |
| 19 | <freezing, cold> |

## 6. DATA ANALYSIS

In this section, I analyze the data collected and discussed in Section 5, specifically as it pertains to the research questions introduced at the beginning of this paper. By exploring patterns in the weak-strong adjective frequency and elements of the adjective relation graph (Figure 1), I will examine the hypothesis that there is a predictable pattern to scalable adjective usage.

## 6.1. Weak-Strong Adjective Frequency

As mentioned in Section 5.3 only eight of 81 identified weak-strong pairs contained two adjectives from the same MPQ category. Of the remaining 73 pairs, many seem to be strongly related, though they are not grouped in the MPQ. Firstly, there are nine pairs of adjectives that are represented more than once in the weak-strong list. Some, like <intense, burning> are found twice in the same configuration, while others can be found ordered in both directions (See 5.3 for example with 'sharp' and 'dull'). In either case, there seems to be some sort of connection between the adjectives such that they are more likely to be used together when describing a painful sensation. Beyond these nine pairs, there are other identified weak-strong pairs with specific relationships between adjectives that are not in the same MPQ category. Specifically, two weak-strong pairs— [sharp, dull] and [cold, hot] —are direct antonyms of one another (as defined by intuition and WordNet [25]). As discussed previously, it is impossible to order antonyms like *cold* and *hot* on a Horn scale such that scalar implicature applies [9],[10]. However, based on the data collected for this research, people sometimes use antonyms with the same constructions that otherwise suggest different gradations of meaning.

Finally, I identified additional weak-strong pairs which are not related by antonymy, but whose meanings are closely related or even synonymous. These pairs include examples like [tingling, prickling] and [aching, throbbing], which are not grouped together in the same MPQ category, but which are intuitively similar. To further explore the different types of relationships between adjectives found in the weak-strong pairs and in the MPQ in general, more robust comparisons using WordNet or survey data would be needed. However, the prevalence of adjectives from different MPQ categories grouped together suggests a need to reevaluate the division of the MPQ categories.

## 6.2. Adjective Graph

The graph labeled Figure 1 (Section 5.3) is a visual representation of the data collected in this research. In this section, I will explore a few key insights from the graph.

***Bidirectional Arrows***. As mentioned in the previous section, there are several adjective node pairs—five in total—which have arrows pointing both from adjective A to adjective B, and from B to A. These bidirectional node pairs show support for the null hypothesis: if an adjective A can be used in both a stronger and weaker position relative to adjective B, that would suggest that the relative intensity of adjectives is not predictable. However, with 22 adjective node pairs, the four bidirectional ones comprise only 23% of the total; the other 77% of adjectives were consistently ordered in relation to their connecting nodes. Interestingly, all the adjectives represented in bidirectional pairings—'numb', 'burning', 'sharp', 'aching', 'dull', 'cold', 'tingling'—are in the top 40% of most frequently occurring MPQ adjectives within weak-strong pairs, and in the top 30% of the corpus overall. Further research could consider whether the frequency of adjectives has an impact on the consistency of usage, and specifically on how people perceive the intensity of high-frequency adjectives.

***Loops.*** In addition to bidirectional arrows, there are also several instances of loops, or adjectives with an edge that starts and ends at the same node. There are eight loops spanning seven adjective nodes: 'tight', 'sore', 'cold', 'intense', 'hurting', 'tingling', 'itching'. These adjectives were each found in a context like the ones given in (25) and (26).

(25)   **itchy but not itchy** to where I am scratching-type feeling
(26)   my muscles are still **tight but not** nearly as **tight** as they used to be

These cases demonstrate that adjectives can not only be part of scales, but themselves have scalable properties. Though the statement 'itchy but not itchy' is (at least, on its face) self-contradictory, the sentence given in (25) compares 'itchy' to 'itchy to where I am scratching-type feeling,' which appears to convey a stronger sensation than just 'itchy.' Further research could attempt to account for the effect of modifiers and predicates on adjective intensity, particularly in contexts where an adjective is compared to a different intensity version of itself.

***Disconnected Nodes.*** Lastly, another starkly apparent visual on the graph is the separation between the 'tight' and 'squeezing' nodes and the rest of the adjectives. While all other adjectives are connected to more than one other adjective node, whether directly or via paths through other nodes, 'tight' points only to 'squeezing' (and itself), and 'squeezing' has no additional emanating arrows. In terms of frequency, both 'tight' and 'squeezing' are in the bottom third of weak-strong pair adjectives, though 'tight' ranks higher up in overall corpus adjective frequency. Again, considering the frequency of certain adjectives in the general lexicon would be an interesting follow-up to this research, and could perhaps address phenomenon such as the isolated [tight, squeezing] pair.

## 7. RESULTS AND DISCUSSION

As demonstrated by the analysis presented in Section 6, there is not a neat, linear relationship between all the adjectives collected. This, however, does not automatically support the null hypothesis $H0_a$, that people do not use adjectives in a predictable and scalable manner. Natural language is dynamic and fluid. So, while the overall picture is somewhat messy, the more informative process is attempting to find patterns within the complexity. Specifically, the graph is considered for its agreement with the intensity scales presented by the MPQ. From the final scales constructed in this research (found in Figure 1 in Section 5.3), there are 17 adjective relationships that can be defined by comparing each element in a given scale to all the other elements in that scale. For example, in the scale <annoying, intense, unbearable>, it is defined not only that *unbearable* is stronger than *intense,* but also that *unbearable* is stronger than *annoying*, etc. These relationships could also be calculated for the same adjectives in the MPQ. Of these 17 adjective relationships, the scales developed in this research demonstrated 58.8% agreement with the MPQ. That is, the HealingWell scales correctly predicted the relative strength of two adjectives as defined by the MPQ 10 out of 17 times. Of the remaining seven, four were incorrectly predicted. For example, where the MPQ defines *hot* as less intense than *burning*, the HealingWell scales defined *hot* as more intense than *burning*. The remaining three were inconclusive, with the HealingWell scales unable to predict the relative intensity between two adjectives. This was the case for the scale <dull/sore/aching, heavy, hurting>, where *dull* and *sore* were both defined as less intense than *heavy*, but no information was obtainable for the intensity of *dull* relative to *sore* and vice versa. Considering only the adjective relationships that the HealingWell scales were able to make any prediction on (i.e., excluding the inconclusive cases), this amounts to a 71.4% agreement with the MPQ.

Since the expected agreement due to chance for these two datasets is 50% (for each two adjectives compared, they were either both in the same order as the MPQ or both in the opposite order), the HealingWell scales demonstrate an above chance level agreement with the MPQ. In analyzing the rate of positive agreement between the HealingWell relationships and the MPQ relationships, there is not statistical significance at the 0.1 alpha level. However, considering the small sample size, looking at only the non-ambiguous relationship predictions might be appropriate. I conducted a one-sample t-test to compare the predictions of the HealingWell scales to the MPQ orderings. With $t(15)=-1.76$, $p=0.098$, and the results show significance at the 0.1 alpha level. At the 0.1 alpha level, the null hypothesis that people do not use adjectives predictably can be rejected. This suggests that there is support for hypothesis H1, that there is some predictability to the way in which people use adjectives to describe their pain.

Even with support for hypothesis H1, it appears that the pattern of scalable adjective use is more nuanced than the MPQ defines it, with the intensity of adjectives not always entirely pinpointable. Furthermore, the HealingWell scales were produced using the categories defined by the MPQ. As discussed above, there is reason to believe that the MPQ categories are not ideally divided based on how people use the MPQ adjectives. Since Horn scales are only reasonable for adjectives that differ in intensity rather than semantic category, further analysis would be needed to consider which adjectival scales should be constructed from the weak-strong pairs collected.

Finally, this research also provided a preliminary investigation into Hypothesis 2, on potential differences in adjective usage between different types of chronic pain sufferers. At the start of this research, I hypothesized that there might be a difference in adjective frequency across different subcategories of chronic pain. In section 4.5, I compared frequency data across different corpora, and between the different topics in the created HealingWell corpus. While RA was an outlier for some adjective frequencies, the differences are not appreciably significant, due to the small sample size for the lower-frequency adjectives (less than 10 or so occurrences in a multi-million-word corpus). Given the overall similarity of adjective frequencies across corpora and forum topics, this research provides preliminary support for $H0_b$, that different categories of chronic pain are not associated with specific adjective pain descriptors.

## 8. CONCLUSION

This study was conducted to test the concepts behind the design of the McGill Pain Questionnaire, a clinical tool for assessing pain quality. Whereas previous work had recreated the MPQ by eliciting survey data, this research attempted to reconstruct the adjective intensity relationships defined by the MPQ by looking only at sentence construction patterns in spontaneous speech. Spontaneous speech more closely approximates the ways in which people understand and use adjectives as pain descriptors. In considering the field of pain assessment, the goal is to facilitate communication of pain from patient to physician. As such, to judge the efficacy of an adjective-based pain questionnaire, it is important to understand how patients describe their pain without prescribed frameworks like the MPQ. The downside to using a natural corpus, like the HealingWell corpus developed in this research, compared to survey data is the unpredictability of the data. Here, very specific patterns of adjective use were required, which were only found in small quantities in the forum corpus. The small sample sizes limit the power of the conclusions drawn in this research. Still, the successful reconstruction of adjective intensity scales from partial orderings, however limited, will hopefully begin an insightful conversation on the structure of current clinical pain questionnaires. For the adjectival pain questionnaire to be a valuable clinical tool, the patient's understanding and use of each adjective must align with the meaning prescribed and interpreted by the receiving physician; in other words, the patient's pain must be communicable through adjectives. This research suggests that, given the right division of adjectives by category, it is possible to predict the relative intensities of adjectives within a given category to some degree. Further research will be needed to determine which categorizations of adjectives are best, and how to find them.

## ETHICAL CONSIDERATIONS

All the collected data comes from HealingWell.com, whose privacy policy states that all forums are public and accessible to guests and search engines [13]. No identifying information—including name, age, gender, or location—was collected in connection to any of the forum data, to protect the privacy and identities of the post authors.

## ACKNOWLEDGEMENTS

I would like to thank my advisor, Professor Fellbaum, for her mentorship and guidance throughout the research and writing process.

I also owe a thank you to my boyfriend Ariel, for his patience and assistance in understanding and optimizing the computer science concepts used for this research.

Finally, I would like to thank my mom for her thoughtful editing and helpful brainstorming.

**Author: Miriam Stern**

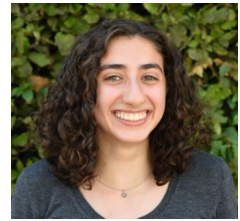

Miriam is an undergraduate student at Princeton University entering her fourth and final year in the Program in Linguistics. She is also completing the pre-med track and will be attending Sydney Kimmel Medical College at Thomas Jefferson University upon completion of her undergraduate studies. Miriam hopes to continue traversing the intersection between language and medicine in her future endeavours.